\documentclass[runningheads]{llncs}

\usepackage{graphicx}

\usepackage{tikz}
\usepackage{comment}
\usepackage{amsmath,amssymb} 
\usepackage{color}

\usepackage[accsupp]{axessibility}  

\usepackage[width=122mm,left=12mm,paperwidth=146mm,height=193mm,top=12mm,paperheight=217mm]{geometry}

\usepackage{multirow}
\usepackage{booktabs}
\usepackage{bm}
\usepackage{algorithm}
\usepackage{algorithmic}

\usepackage{setspace}
\def\eqref#1{Equation~\ref{#1}}
\def\figref#1{Figure~\ref{#1}}
\def\tableref#1{Table~\ref{#1}}
\def\algref#1{Algorithm~\ref{#1}}

\def\vtheta{{\bm{\theta}}}
\def\vzero{{\bm{0}}}
\def\vg{{\bm{g}}}
\def\vh{{\bm{h}}}

\def\vw{{\bm{w}}}
\def\vx{{\bm{x}}}

\def\gP{{\mathcal{P}}}
\DeclareMathOperator*{\argmin}{arg\,min}
\newcommand*\mean[1]{\bar{#1}}

\usepackage{xspace}
\newcommand{\fl}{\textsc{Federated Learning}\xspace}

\newcommand{\fedavg}{\textsc{FedAvg}\xspace}

\newcommand{\scaffold}{\textsc{SCAFFOLD}\xspace}
\newcommand{\fedprox}{\textsc{FedProx}\xspace}
\newcommand{\fedsplit}{\textsc{FedSplit}\xspace}
\newcommand{\fedpd}{\textsc{FedPd}\xspace}
\newcommand{\feddyn}{\textsc{FedDyn}\xspace}
\newcommand{\ours}{\textsc{AdaBest}\xspace}

\newcommand{\vrlsgd}{\textsc{VRL-SGD}\xspace}
\newcommand{\fsvrg}{\textsc{FSVRG}\xspace}
\newcommand{\dane}{\textsc{DANE}\xspace}
\newcommand{\aide}{\textsc{AIDE}\xspace}
\newcommand{\feddane}{\textsc{FedDANE}\xspace}

\newcommand{\localsgd}{\textsc{LocalSGD}\xspace}
\newcommand{\sgd}{\textsc{SGD}\xspace}
\newcommand{\rvsgd}{\textsc{RV-SGD}\xspace}
\newcommand{\rvlsgd}{\textsc{RV-LSGD}\xspace}
\newcommand{\svrg}{\textsc{SVRG}\xspace}
\newcommand{\sarah}{\textsc{SARAH}\xspace}

\def\cifart{{\textsc{CIFAR10}}}
\def\cifarh{{\textsc{CIFAR100}}}
\def\emnist{{\textsc{EMNIST-L}}}

\begin{document}

\pagestyle{headings}
\mainmatter

\title{AdaBest: Minimizing Client Drift in Federated Learning via Adaptive Bias Estimation} 

\titlerunning{AdaBest: Minimizing Client Drift in Federated Learning}
%
\author{Farshid Varno\inst{1,2}
\and
Marzie Saghayi\inst{1}
\and
Laya Rafiee Sevyeri\inst{2,3}
\and
Sharut Gupta\inst{2,4}
\and
Stan Matwin\inst{1,5}
\and
Mohammad Havaei\inst{2}
}
%
\authorrunning{Published as a conference paper at ECCV 2022 - some corrections applied}
%
\institute{
Dalhousie University, Halifax, Canada\\
\email{\{f.varno,m.saghayi\}@dal.ca}, \ \email{stan@cs.dal.ca}
\and
Imagia Cybernetics Inc., Montreal, Canada
\email{\{laya.rafiee,sharut.gupta,mohammad.havaei\}@gmail.com}
\and
Concordia University, Montreal, Canada
\and
Indian Institute of Technology Delhi, New Delhi, India
\and
Polish Academy of Sciences, Warsaw, Poland
}
\maketitle

\begin{abstract}

In Federated Learning (FL), a number of clients or devices collaborate to train a model without sharing their data. Models are optimized locally at each client and further communicated to a central hub for aggregation. While FL is an appealing decentralized training paradigm, heterogeneity among data from different clients can cause the local optimization to {\it drift} away from the global objective. 
In order to estimate and therefore remove this drift, variance reduction techniques have been incorporated into FL optimization recently. However, these approaches inaccurately estimate the clients' drift and ultimately fail to remove it properly. 
In this work, we propose an adaptive algorithm that accurately estimates drift across clients. 
In comparison to previous works, our approach necessitates less storage and communication bandwidth, as well as lower compute costs. Additionally, our proposed methodology induces stability by constraining the norm of estimates for client drift, making it more practical for large scale FL. Experimental findings demonstrate that the proposed algorithm converges significantly faster and achieves higher accuracy than the baselines across various FL benchmarks.

\end{abstract}

\section{Introduction}
In Federated Learning (FL), multiple sites with data often known as \emph{clients} collaborate to train a model by communicating parameters through a central hub called \emph{server}. 
At each round, the server broadcasts a set of model parameters to a number of clients.
Selected clients separately optimize towards their local objective.
The locally trained parameters are sent back to the server, where they are aggregated to form a new set of parameters for the next round.
A well-known aggregation is to simply average the parameters received from the participating clients in each round.
 This method is known as 
\fedavg \cite{mcmahan2017communication} or \localsgd~\cite{stich2018local}.

In order to reduce the communication costs as well as privacy concerns (for example possible leakage of data from gradients \cite{zhu2020deep}), multiple local optimization steps are often preferable and sometimes inevitable \cite{mcmahan2017communication}.
Unfortunately, multiple local updates subject the parameters to \emph{client drift} \cite{karimireddy2020scaffold}. While \sgd is an unbiased gradient descent estimator,
\localsgd is biased due to the existence of client drift.
As a result, \localsgd converges to a neighborhood around the optimal solution with a distance proportionate to the magnitude of the bias \cite{ajalloeian2020convergence}. 
The amount of this bias itself depends on the heterogeneity among the clients' data distribution, causing \localsgd to perform poorly on non-iid benchmarks \cite{zhao2018federated}.

One effective way of reducing client drift is by adapting Reduced Variance SGD (\rvsgd) methods \cite{johnson2013accelerating,roux2012stochastic,shalev2013stochastic,nguyen2017sarah} to \localsgd.
The general strategy is to regularize the local updates with an estimate of gradients of inaccessible training samples (i.e., data of other clients).
In other words, the optimization direction of a client is modified using the estimated optimization direction of other clients.
These complementary gradients could be found for each client $i$ by subtracting an estimate of the local gradients from an estimate of the oracle's\footnote{Oracle dataset refers to the hypothetical dataset formed by stacking all clients' data.
Oracle gradients are the full-batch gradients of the Oracle dataset.} full gradients.
In this paper, we refer to these two estimates with $\vh_i$ and $\vh$, respectively.
Therefore, a reduced variance local gradient for client $i$ would be in general form of $\nabla L_i + (\vh - \vh_i)$ where $\nabla L_i$ corresponds to the true gradients of the local objective for client $i$.

The majority of existing research work on adapting \rvsgd to distributed learning do not meet the requirement to be applied to FL.
Some proposed algorithms require full participation of clients \cite{shamir2014communication,reddi2016aide,liang2019variance}; thus, are not scalable to \emph{cross-device} FL\footnote{In contrast to \emph{cross-silo} FL, cross-device FL is referred to a large-scale (in terms of number of clients) setting in which clients are devices such as smart-phones.}.
Another group of algorithms require communicating the true gradients \cite{li2019feddane,murata2021bias} and, as a result, completely undermine the FL privacy concerns such as attacks to retrieve data from true gradients \cite{zhu2020deep}.

\scaffold \cite{karimireddy2020scaffold} is an algorithm that supports partial participation of clients and does not require the true gradients at the server.
While \scaffold shows superiority in performance and convergence rate compared to its baselines, it consumes twice as much bandwidth. 
To construct the complementary gradients, it computes and communicates $\vh$ as an extra set of parameters to each client along with the model parameters.
\feddyn \cite{acar2021federated} proposed to apply $\vh$ in a single step prior to applying any local update, and practically found better performance and convergence speed compared to \scaffold.
Since applying $\vh$ uses the same operation in all participating clients, \cite{acar2021federated} moved it to the server instead of applying on each client.
This led to large savings of local computation, and more importantly to use the same amount of communication bandwidth as vanilla \localsgd (i.e., \fedavg), which is half of what \scaffold uses.

\feddyn make several assumptions that are often not satisfied in large-scale FL.
These assumptions include having prior knowledge about the total number of clients, a high rate of re-sampling clients, and drawing clients uniformly from a stationary set.
Even with these assumptions, we show that $\vh$ in \feddyn is pruned to explosion, especially in large-scale setting. 
This hurts the performance and holds the optimization back from converging to a stationary point.

\begin{figure}[t]
    \centering
    \def\svgwidth{12.0cm}
    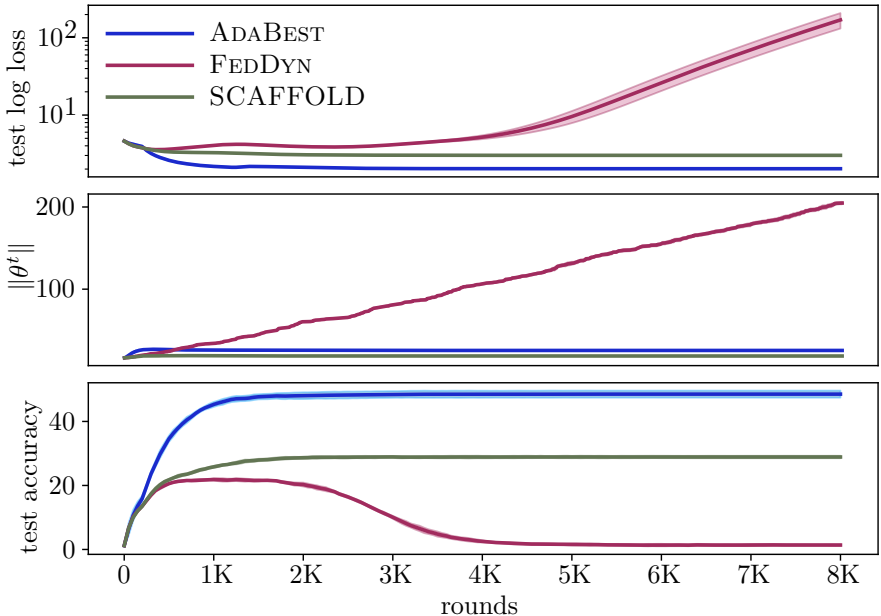
    \caption{Asymptotic instability of \feddyn as a results of unbounded increase of $\|\vh^t\|$. From top to bottom test loss (log scale), norm of cloud parameters, and test accuracy are shown in subplots. The shared horizontal axis shows the number of communication rounds. Each experiment is repeated 5 times with different random seed of data partitioning. Solid lines and shades represent mean and standard deviation respectively}
    \label{fig:stability_c100}
\end{figure}

This paper proposes \ours, a Reduced Variance LocalSGD (\rvlsgd) solution to minimize the client drift in FL. Compared to the baselines, we define a simpler yet more elegant method of incorporating previous observations into estimates of complementary gradients. Our solution alleviates the instability of the norm of $\vh$ in \feddyn (see \figref{fig:stability_c100} for empirical evaluation) while consuming the same order of storage and communication bandwidth, and even reducing the compute cost (see supplementary material for quantitative comparison). 
Unlike previous works, our algorithm provides a mechanism for adapting to changes in the distribution of client sampling and does not require prior knowledge of the entire set of clients.
These characteristics of \ours, combined with its stability, provide a practical solution for large-scale cross-device FL. Our main contributions are as follows:

\begin{itemize}
    \item We show that the existing \rvlsgd approaches for cross-device FL fail to efficiently converge to a stationary point. In particular, the norm of the parameters in \feddyn is pruned to explosion (see section \ref{sec:compare} for discussion).
    \item We formulate a novel arithmetic approach for implicit accumulation of previous observations into the estimates of oracle full gradients ($\vh$).
    \item Using the new formulation, we present \ours, a novel algorithm that can be thought as a generalization of both \fedavg and \feddyn. We introduce a new factor $\beta$ that stabilizes our algorithm through controlling the norm of $\vh$. As a result, the optimization algorithm converges to a stationary point (see Sections \ref{sec:compare} for detailed discussion). Unlike baselines, \ours does not assume that the set of training clients are stationary nor it requires a prior knowledge about its cardinality.
    \item We conduct various experiments under different settings of number of clients, balance among partitions, and heterogeneity. 
    Our results indicate superior performance of \ours in nearly all benchmarks 
    in addition to significant improvements in stability and convergence rate. 
    In benchmarks with a large number of clients, our method outperforms the baselines by a wide margin. 
    The results show nearly a two-fold improvement in test accuracy over the second-best candidate in some cases.
\end{itemize}

\section{Related Work}
\label{subsec:related}

A major challenge in FL is data heterogeneity across clients where the local optima in the parameter space at each client may be far from that of the global optima. This causes a {\it drift} in the local parameter updates with respect to the server aggregated parameters. Recent research has shown that in such heterogeneous settings, \fedavg is highly pruned to client drift \cite{zhao2018federated}. 

To improve the performance of FL with heterogeneous data, some previous works use knowledge distillation to learn the cloud model from an ensemble of client models. This approach has been shown to be more effective than simple parameter averaging in reducing bias of the local gradients \cite{lin2020ensemble,li2019fedmd,zhu2021data}.

Another group of methods can be categorized as {\it gradient based} in which the gradients are explicitly constrained on the clients or server for bias removal. \fedprox \cite{li2020federated} penalizes the distance between the local and cloud parameters whereas \cite{wang2020tackling} normalizes the client gradients prior to aggregation. Inspired by momentum SGD, \cite{yu2019linear} uses a local buffer to accumulate gradients from previous rounds at each client and communicate the momentum buffer with the server as well as the local parameters which doubles the consumption of communication bandwidth. Instead of applying momentum on the client level, \cite{hsu2019measuring} and \cite{wang2019slowmo} implement a server momentum approach which avoids increasing communication costs. 

Inspired by Stochastic Variance Reduction Gradients (\svrg) \cite{johnson2013accelerating}, some works incorporate variance reduction into local-SGD \cite{acar2021federated,li2019feddane,karimireddy2020scaffold,liang2019variance,zhang2020fedpd,konevcny2016federated,murata2021bias,nguyen2017sarah}. \dane\cite{shamir2014communication}, \aide\cite{reddi2016aide}, and \vrlsgd\cite{liang2019variance} incorporated \rvsgd in distributed learning for full client participation. \feddane\cite{li2019feddane} is an attempt to adapt \dane to FL setting\footnote{Recall that Federated Learning is a sub-branch of distributed learning with specific characteristics geared towards practicality\cite{mcmahan2017communication}.}, though it still undermines the privacy concerns such as attacks to retrieve data from true gradients \cite{zhu2020deep}. 
Different from our work, most methods in this category such as \vrlsgd \cite{liang2019variance}, \fsvrg \cite{konevcny2016federated}, \fedsplit \cite{pathak2020fedsplit} and \fedpd \cite{zhang2020fedpd} require full participation of clients which makes them less suitable for cross-device setting where only a fraction of clients participate in training at each round. While \feddane \cite{li2019feddane} works in partial participation, empirical results show it performs worse than Federated Averaging \cite{acar2021federated}. 
More comparable to our method are those capable of learning in partial participation setting. In particular, \scaffold \cite{karimireddy2020scaffold} uses control variates on both the server and clients to reduce the variance in local updates. In addition to the model parameters, the control variates are also learned and are communicated between the server and the clients which take up additional bandwidth. \cite{murata2021bias} also reduces local variance by estimating the local bias at each client and using an \svrg-like approach to reduce the drift.
While \scaffold applied variance reduction on clients, \feddyn \cite{acar2021federated} applies it partly on the server and partly on the clients.  
Our proposed method, is probably closer to \feddyn than to others; however, they differ in the way gradients are estimated. See section \ref{sec:compare} for detailed comparison of \ours with \feddyn and \scaffold.


\section{Method}

In this section, we present an overview of the general FL setup and further introduce our notation to formulate the problem statement. Next we detail the proposed algorithm and how to apply it. Finally, we demonstrate the efficacy of our technique by comparing it with the most closely related approaches.

\subsection{Federated Learning}
We assume a standard FL setup in which a central server communicates parameters to a number of clients.
The goal is to find an optimal point in the parameter space that solves a particular task while clients keep their data privately on their devices during the whole learning process. 

Let $S^t$ be the set of all registered clients at round $t$ and $\gP^t$ be a subset of it drawn from a distribution $P(S^{\tau}; \tau=t)$.
The server broadcasts the \emph{cloud model} $\vtheta^{t-1}$ to all the selected clients.
Each client $i \in \gP^t$, optimizes the cloud model based on its local objective and transmits the optimized \emph{client model}, $\vtheta_i^{t}$ back to the server.
The server aggregates the received parameters and prepares a new cloud model for the next round. 
\tableref{tb:notation} lists the most frequently used symbols in this paper along with their meanings. Note that the \emph{aggregate model} is the average of \emph{client models} over values of $i \in \gP^t$.
Likewise, aggregate gradients are the average of the client gradients.

\begin{table}
\caption{Summary of notion used in this paper}
\label{tb:notation}
    \begin{center}
    \begin{tabular}{cl}
        \toprule
        $u^t, \ u_i^t, \ u_i^{t,\tau}$ & \textbf{variable} $u$ at \{round $t$, and {client} $i$, and {local step} $\tau$\} \\
        $|\ . \ |, \ \langle \ . \ , \ . \ \rangle, \ u^{(v)}$ \ \ \ \ \ \ &  \textbf{cardinality},\  \textbf{inner product}, \ \textbf{power}\\
        $\ {\| . \|}^2, \ \angle(. , .)$ & \textbf{2-norm squared}, \textbf{angle} \\
        $S^t, \ \gP^t$ & set of $\{{\text{all}},\text{round}\}$ \textbf{clients}\\
        $\vtheta^t,\  \bar{\vtheta}^t, \ \vtheta_i^t,\ \vtheta_i^{t,\tau}$ & $\{\text{cloud},\text{aggregate},\text{client},\text{local}\}$ \textbf{model}\\
        $\vg^t, \ \bar{\vg}^t,  \ \vg_i^t, \ \vg_i^{t,\tau}$ & $\{\text{oracle},\text{aggregate},\text{client},\text{local}\}$ \textbf{gradients}\\
        $\vh^t, \ \vh_i^t$ & $\{\text{full},\text{client}\}$ \textbf{gradients estimates}\\
        \bottomrule
    \end{tabular}
\end{center}
\end{table}

\subsection{Adaptive Bias Estimation} 
Upon receiving the client models of round $t$ ($\{\forall i \in \gP^t: \vtheta_i^t\}$) on the server, the {aggregate model}, $\bar{\vtheta}^t$ is computed by averaging them out. 

{\definition Pseudo-gradient of a variable $u$ at round $t$ is $u^{t-1}-u^{t}$.} 

\begin{remark}
\label{re:aggregate}
Aggregating client models by averaging is equivalent to applying a gradient step of size 1 from the previous round's cloud model using average of client pseudo-gradients or mathematically it is $\bar{\vtheta}^t \leftarrow \frac{1}{|S^t|} \sum_{i\in S^t} \vtheta_i^t = \vtheta^{t-1} - \bar{\vg}^t$.
\end{remark}

\begin{figure}
\centering
\begin{tikzpicture}[scale=0.8]
\draw[fill] (0,0.5) circle [radius=0.05];
\node [red, above] at (0,0.5) {$\bar{\vtheta}^{t-1}$};
\draw [->] [thick] (0,0.5) -- (1.75,1.5);

\node [below right] at (0.6,1.2) {$-\vh^{t-1}$};

\draw[fill] (1.8,1.5) circle [radius=0.05];
\node [above] at (1.8,1.5) {${\vtheta}^{t-1}$};
\draw [->] [thick, red] (1.8,1.5) -- (4.45,0.55);
\node [red] at (3,1.3) {$-\bar{\vg}^{t}$};

\draw [dashed] [thick, blue] (0,0.5) -- (4.5,0.5);
\node [blue] at (2.5,0.2) {$-\frac{1}{\beta}\vh^{t}$};

\draw[fill] (4.5,0.5) circle [radius=0.05];
\node [red, above right] at (4.5,0.5) {$\bar{\vtheta}^{t}$};
\draw [->] [thick] (4.5,0.5) -- (6.45,0.5);
\node at (5.6,0.2) {$-\vh^{t}$};

\draw[fill] (6.5,0.5) circle [radius=0.05];
\node [above right] at (6.5,0.5) {${\vtheta}^{t}$};
\end{tikzpicture}
\caption{Geometric interpretation of \ours's correction applied to the server updates. Server moves the aggregate parameters in the direction of $\mean{\vtheta}^{t-1} - \mean{\vtheta}^{t}$ before sending the models to the next round's clients}
\label{fig:geometric_int}
\end{figure}
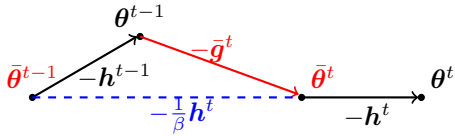

\noindent Next, the server finds the \emph{cloud model} ${\vtheta}^t$ by applying the estimate of the oracle gradients $\vh^t$; that is 
\begin{equation}
    \vtheta^t \leftarrow \bar{\vtheta}^t - \vh^t,
    \label{eq:theta}
\end{equation}
where $\vh^t$ is found as follows
\begin{equation}
    \vh^t=\beta(\bar{\vtheta}^{t-1}-\bar{\vtheta}^t).
    \label{eq:gt}
\end{equation}
In section \ref{sec:adaptability}, we further discuss the criteria for chosen $\beta$ which leads to a fast convergence.
The described cycle continues by sending the cloud model to the clients sampled for the next round ($t+1$) while the aggregate model ($\bar{\vtheta}^t$) is retained on the server to be used in calculation of $\vh^{t+1}$ or deployment.
A schematic of the geometric interpretation of the additional drift removal step taken at the server is shown in \figref{fig:geometric_int}.

After receiving the cloud model, each client $i \in \gP^t$, optimizes its own copy of the model towards its local objective, during which the drift in the local optimization steps is reduced using the client's pseudo-gradients stored from the previous rounds (see \algref{alg:adabest}). 
The modified client objective is $\argmin_{\vtheta} \mathfrak{R}_i(\vtheta^t)$ such that 
$\mathfrak{R}_i(\vtheta^t) = L_i(\vtheta^t) - \mu \langle \vtheta^t, \vh_i^{t_i'} \rangle, \ $ where $L_i$ is the local empirical risk defined by the task and data accessible by the client $i$, and $t_i'$ is the last round that client $i$ participated in the training.
Accordingly, the local updates with step size $\eta$ becomes
\begin{equation}
    \vtheta_i^{t,\tau} \leftarrow \vtheta_i^{t,\tau-1} - \eta(\nabla L_i(\vtheta_i^{t,\tau-1})-\mu \vh_i^{t_i'}),
    \label{eq:theta_i_tau}
\end{equation}
where $\mu$ is the \emph{regularization factor} (\feddyn has a similar factor; see supplementary material for further discussion on the choice of optimal value for $\mu$).

After the last local optimization step, each sampled client updates the estimate of its own local gradients and stores it locally to be used in the future rounds that the client participate in the training.
This update is equivalent to $\vh_i^t=\frac{1}{t-t_i'} \vh^{t_i'} + \vg_i^{t}$ where $\vg_i^{t} = \vtheta^{t-1}-\vtheta_i^t$. Finally, the participating clients send the optimized model $\vtheta_i^t$ back to the server. Our method along with \scaffold and \feddyn is presented in \algref{alg:adabest}.

\begin{algorithm}
   \caption{\colorbox{red!30}{\strut\scaffold/m}, \colorbox{blue!20}{\strut\feddyn}, and \colorbox{green!30}{\strut\ours}}
   \label{alg:adabest}
\begin{spacing}{1.2}
\begin{algorithmic}
   \STATE {\bfseries Input:} $T, \vtheta^0, \mu, \beta$
   \FOR{$t=1$ {\bfseries to} $T$}
   \STATE ~Sample clients $\gP^t \subseteq S^t$.
   \STATE ~Transmit $\vtheta^{t-1}$ to each client in $\gP^t$
   \STATE \colorbox{red!30}{Transmit $h^{t-1}$ to each client in $\gP^t$ (\scaffold/m)}
   \FOR{each client $i \in \gP^t$ {\bfseries in parallel}} 
   \STATE \COMMENT{receive cloud model}
   \STATE $\vtheta_i^{t,0} \leftarrow \vtheta^{t-1}$
   \STATE \COMMENT{locally optimize for $K$ local steps}
   \FOR{$k=1$ {\bfseries to} $K$}
   \STATE Compute mini-batch gradients $L_i(\vtheta_i^{t,k-1})$
   \STATE \colorbox{red!30}{$\vg_i^{t,k-1} \leftarrow \nabla L_i(\vtheta_i^{t,k-1}) - \vh_i^{t'_i} + \vh^t$ (\scaffold/m)}
   \STATE \colorbox{blue!20}{$\vg_i^{t,k-1} \leftarrow \nabla L_i(\vtheta_i^{t,k-1}) - \vh_i^{t'_i} - \mu (\vtheta^{t-1} - \vtheta_i^{t, k-1})$ (\feddyn)}
   \STATE \colorbox{green!30}{$\vg_i^{t,k-1} \leftarrow \nabla L_i(\vtheta_i^{t,k-1}) - \vh_i^{t'_i}$ (\ours)}
  \STATE ~$\vtheta_i^{t,k} \leftarrow \vtheta_i^{t,k-1} - \eta \vg_i^{t,k-1}$
   \ENDFOR
   \STATE \COMMENT{update local gradient estimates}
   \STATE ~$\vg_i^{t} \leftarrow \vtheta^{t-1} - \vtheta_i^{t,K}$
    \STATE \colorbox{red!30}{$\vh_i^{t} \leftarrow \vh_i^{t'} - \vh^{t-1} + \frac{1}{K\eta} \vg_i^{t}$
    (\scaffold/m)}
   \STATE \colorbox{blue!20}{$\vh_i^{t} \leftarrow \vh_i^{t'_i} + \mu \vg_i^{t}$ (\feddyn)}
   \colorbox{green!30}{$\vh_i^{t} \leftarrow \frac{1}{t-t'_i}\vh_i^{t'_i} + \mu \vg_i^{t}$ (\ours)}
   \STATE ~$t'_i \leftarrow t$
   \STATE ~Transmit client model $\vtheta_i^t:=\vtheta_i^{t,K}$.
   \ENDFOR
   \STATE \COMMENT{aggregate received models}
   \STATE ~$\mean{\vtheta^t} \leftarrow \frac{1}{|\gP^t|}\sum_{i\in \gP^t}\vtheta_i^t$
   \STATE \COMMENT{update oracle gradient estimates}
   \STATE \colorbox{red!30}{$\vh^{t} \leftarrow \frac{|S^t|-|\gP^t|}{|S^t|}\vh^{t-1} +\frac{|\gP^t|}{K\eta|S^t|}({\vtheta}^{t-1}-\mean{\vtheta}^t)$ (\scaffold/m)}
   \STATE \colorbox{blue!20}{$\vh^{t} \leftarrow \vh^{t-1} +\frac{|\gP^t|}{|S^t|}({\vtheta}^{t-1}-\mean{\vtheta}^t)$ (\feddyn)}
   \colorbox{green!30}{$\vh^t \leftarrow \beta(\mean{\vtheta}^{t-1}-\mean{\vtheta}^t)$ (\ours)}
   \STATE \COMMENT{update cloud model}
  \STATE \colorbox{red!30} {$\vtheta^{t} \leftarrow \mean{\vtheta}^{t}$ (\scaffold/m)}
  \STATE \colorbox{blue!20} {$\vtheta^{t} \leftarrow \mean{\vtheta}^{t} - \vh^t$ (\feddyn)}
  \colorbox{green!30} {$\vtheta^{t} \leftarrow \mean{\vtheta}^{t} - \vh^t$ (\ours)}
   \ENDFOR
\end{algorithmic}
\end{spacing}
\end{algorithm}

\subsection{Relation to \rvsgd}
Stochastic Variance Reduction Gradients (\svrg) \cite{johnson2013accelerating} and its variants\cite{nguyen2017sarah,bi2021variance,konevcny2014ms2gd,babanezhad2015stopwasting,xiao2014proximal} are of the most recent and popular \rvsgd algorithms. Given parameters $\vw$ and an objective function $\ell$ to minimize it modifies \sgd update from $\vw^k \leftarrow \vw^{k-1} - \eta \nabla \ell (\vw^{k-1}, \vx_k)$ to
\begin{equation}
    \vw^k \leftarrow \vw^{k-1}- \eta (\nabla \ell (\vw^{t-1}, \vx_k) + \mathcal{G}(\tilde{\vw})-\ell(\tilde{\vw}, \vx_k)),
\end{equation}
where $\vx_k$ is a sample of data, $\tilde{\vw}$ is a snapshot of $\vw$ in the past and $\mathcal{G} (\tilde{\vw})$ is full batch gradients at $\tilde{\vw}$. The analytic result of this unbiased modification is that if the empirical risk is strongly convex and the risk function over individual samples are both convex and L-smooth  then the error in estimating gradients of $\mathcal{G} (\tilde{\vw})$ is not only bounded but also linearly converges to zero (refer to \cite{johnson2013accelerating} for proof). Under some conditions, \cite{babanezhad2015stopwasting} showed that this convergence rate is not largely impacted if a noisier estimate than original $\mathcal{G} (\tilde{\vw})$ proposed by \svrg is chosen for full batch gradients. \cite{bi2021variance} investigated applying \svrg in non-convex problems and \cite{konevcny2014ms2gd} generalized it for mini-batch \sgd. \cite{nguyen2017sarah} proposed \sarah, a biased version of \svrg that progressively updates the estimate for full gradients for optimizations steps applied in between taking two snapshots. 
Our algorithm could be thought as a distributed variant of \sarah where 
\begin{enumerate}
    \item $\mathcal{G} (\tilde{\vw})$ is approximated by biased pseudo-gradients (and renamed to $\vh$).
    \item The outer loop for taking the snapshot is flattened using an exponential weighted average.
\end{enumerate}    


\subsection{Relation to FL Baselines}
\label{sec:compare}
\algref{alg:adabest} demonstrates where our method differs from the baselines by color codes.
Compared to the original \scaffold, we made a slight modification in the way communication to the server occurs, preserving a quarter of the communication bandwidth usage.
We refer to this modified version as \scaffold/m.
In the rest of this section, we will discuss the key similarities and differences between our algorithm, \feddyn and \scaffold in terms of cost, robustness and functionality.

\subsubsection{Cost}
\scaffold consumes twice as much communication bandwidth as \feddyn and \ours. This should be taken into account when comparing the experimental performance and convergence rate of these algorithms. All of these three algorithms require the same amount of storage on the server and on each client. Finally, \ours has a lower compute cost compared to \feddyn, \scaffold both locally (on clients) and on the server. We provide quantitative comparison of these costs in supplementary material.

\subsubsection{Robustness}
According to the definition of cross-device FL, the number of devices could even exceed the number of examples per each device \cite{mcmahan2017communication}. In such a massive pool of devices, if the participating devices are drawn randomly at uniform (which our baselines premised upon), there is a small chance for a client to be sampled multiple times in a short period of time. In \feddyn, however, $\vh^t=\sum_{\tau=1}^t\frac{|\gP^t|}{|S^t|}\mean{\vg}^t$, making it difficult for the norm of $\vh$ to decrease if pseudo-gradients in different rounds are not negatively correlated with each other (see Theorem \ref{re:feddyn_decrease_norm} and its proof in supplementary material). In case clients are not re-sampled with a high rate then this negative correlation is unlikely to occur due to changes made to the state of the parameters and so the followup pseudo-gradients (see Section \ref{sec:adaptability} for detailed discussion). A large norm of $\vh^t$ leads to a large norm of $\vtheta^{t}$ and in turn a large ${\|\mean{\vtheta}^{t+1}\|}^2$. This process is exacerbated during training and eventually leads to exploding norm of the parameters (see \figref{fig:stability_c100}). 
In Section \ref{sec:adaptability}, we intuitively justify the robustness of \ours for various scale and distribution of client sampling.

\begin{theorem}
\label{re:feddyn_decrease_norm}
In \feddyn, ${\|\vh^t\|}^2 \leq {\|\vh^{t-1}\|}^2$ requires
\begin{equation}
\cos( \angle (\vh^{t-1}, \mean{\vg}^{t})) \leq - \frac{|\gP^t|}{2|S^t|} \frac{\|\mean{\vg}^{t}\|}{\|{\vh}^{t-1}\|}.
\end{equation}

\end{theorem}

\subsubsection{Functionality}
\ours allows to control how far to look back through the previous rounds for effective estimation of full and local gradients compared to existing \rvlsgd baselines.
To update the local gradient estimates, we dynamically scale the previous values down because the period between computing and using $\vh_i^{t_i'}$ on client $i$ (the period between two participation, i.e., $t-t_i'$) can be long during which the error of estimation may notable increase.
See \algref{alg:adabest} for comparing our updates on local gradients estimation compared to that of the baselines.
Furthermore, at the server, $\vh^t$ is calculated as the weighted difference of two consecutive aggregate models.
Note that, if expanded as a power series, this difference by itself is equivalent to accumulating pseudo-gradients across previous rounds with an exponentially weighted factor.
This series is presented in Remark \ref{re:gt_pow} for which the proof is provided in the supplementary material.
Unlike previous works, proposed pseudo-gradients' accumulation does not necessitate any additional effort to explicitly maintain information about the previous rounds.
Additionally, it reduces the compute cost as quantitatively shown in the supplementary material.
It is a general arithmetic; therefore, could be adapted to work with our baselines as well.

\begin{remark}
\label{re:gt_rec}
$\mean{\vtheta}^{t-1}-\mean{\vtheta}^{t}$ is equivalent to $\vh^{t-1} + \mean{\vg}^{t}$ in \ours.
\end{remark}

\begin{remark}
\label{re:gt_pow}
Cloud pseudo-gradients of \ours form a power series of $\vh^t = \sum_{\tau=1}^{t} \beta^{(t-\tau)}\mean{\vg}^t$, given that superscript in parenthesis means power.
\end{remark}

\subsection{Adaptability} 
\label{sec:adaptability}
As indicated earlier, the error in estimation of oracle full gradients in \feddyn is only supposed to be eliminated by using pseudo-gradients. A difficult learning task, both in terms of optimization and heterogeneity results in a higher variance of pseudo-gradients when accompanies with a low rate of client participation. The outcome of constructing a naive estimator by accumulating these pseudo-gradients is sever. This is shown in \figref{fig:stability_c100}, where on average, there is a long wait time between client re-samples due to the large number of participating clients.
The results of this experiment empirically validates that ${\|\vtheta^t\|}^2$ in \feddyn grows more rapidly and to a much higher value than \ours.
\scaffold is prune to the same type of failure; however, because it scales down previous values of $\vh$ in its accumulation, the outcomes are less severe than that of \feddyn.
In the supplementary material, we present, similar analysis, for a much simpler task (classification on \emnist{}).
It is important not to confuse the source of \feddyn's instability with overfitting (see supplementary material for overfitting analysis). 
However, our observations imply that the stability of \feddyn decreases with the difficulty of the task.

Our parameter $\beta$ solves the previously mentioned problem with norm of $\vh$.
It is a scalar values between 0 and 1 that acts as a slider knob to determine the trade-off between the amount of information maintained from the previous estimate of full gradient and the estimation that a new round provides. On an intuitive level, a smaller $\beta$ is better to be chosen for a more difficult task (both in terms of optimization and heterogeneity) and lower level participation--and correspondingly higher round to round variance among pseudo-gradients and vice versa. We provide an experimental analysis of $\beta$ in the supplementary material; however, in a practical engineering level, $\beta$ could be dynamically adjusted based on the variance of the pseudo-gradients. The goal of this paper is rather showing the impact of using $\beta$. Therefore, we tune it like other hyper-parameters in our experiments.
We leave further automation for finding an optimal $\beta$ to be an open question for the future works.

\begin{remark}
\label{re:spectrum_fedavg}
\fedavg is a special case of \ours where $\beta=\mu=0$.
\end{remark}

\begin{remark}
\label{re:spectrum}
Server update of \feddyn is a special case of \ours where $\beta=1$ except that \textbf{an extra $\frac{|\gP|}{|S|}$ scalar is applied which also adversely makes \feddyn require prior knowledge about the number of clients}.
\end{remark}

\begin{theorem}
\label{re:converge_cond}
If $S$ be a fixed set of clients, $\mean{\vtheta}$ does not converge to a stationary point unless $\vh \rightarrow 0$.
\end{theorem}
As mentioned in Section \ref{sec:compare} and more particular with Theorem \ref{re:feddyn_decrease_norm}, \feddyn is only able to decrease norm of $\vh$ if pseudo-gradients are negatively correlated with oracle gradient estimates which could be likely only if the rate of client re-sampling is high. Therefore, with these conditions often being not true in large-scale FL and partial-participation, it struggles to converge to an optimal point. \scaffold has a weighting factor that eventually could decrease $\|\vh\|$ but it is not controllable. Our algorithm enables a direct decay of $\vh$ through decaying $\beta$. We apply this decay in our experiments when norm of $\vh$ plateaus (see Section \ref{sec:experiments}). This is consistent with Theorem \ref{re:converge_cond} which states that converging to a stationary point require $\vh \rightarrow 0$.

\section{Experiments}

\subsection{Setup}
We evaluate performance and speed of convergence of our algorithm against state-of-the-art baselines. We concentrate on FL classification tasks defined using three well-known datasets. These datasets are, the letters classification task of \emnist{} \cite{lecun1998gradient} for an easy task, \cifart{} \cite{krizhevsky2009learning} for a moderate task and \cifarh{} \cite{krizhevsky2009learning} for a challenging task. The training split of the dataset is partitioned randomly into a predetermined number of clients, for each task. 10\% of these clients are set aside as validation clients and only used for evaluating the performance of the models during hyper-parameter tuning. The remaining 90\% is dedicated to training. The entire test split of each dataset is used to evaluate and compare the performance of each model. Our assumption throughout the experiments is that, test dataset, oracle dataset, and collective data of validations clients have the same underlying distribution. 

To ensure consistency with previous works, we follow \cite{acar2021federated} to control heterogeneity and sample balance among client data splits. For  heterogeneity, we evaluate algorithms in three modes: IID, $\alpha=0.3$ and $\alpha=0.03$. The first mode corresponds to data partitions (clients' data) with equal class probabilities. For the second and third modes, we draw the skew in each client's labels from a Dirichlet distribution with a concentration parameter $\alpha$.
For testing against balance of sample number, we have two modes: \emph{balanced} and \emph{unbalanced} such that in the latter, the number of samples for each client is sampled from a log-normal distribution with concentration parameter equal to $0.3$. 

Throughout the experiments we consistently keep the local learning rate, number of local epochs and batch size as $0.1$, $5$ and $45$ respectively. Local learning rate is decayed with a factor of 0.998 at each round. As tuned by \cite{acar2021federated}, the local optimizer uses a weight decay of $10^{-4}$ for the experiments on \emnist{} and $10^{-3}$ for the experiment on \cifart{} and \cifarh{}. Further details about the optimization is provided in supplementary material.

To tune the hyper-parameters we first launch each experiment for 500 rounds.
$\mu$ of \feddyn is chosen from $[0.002, 0.02, 0.2]$, with $0.02$ performing best in all cases except \emnist, where $0.2$ also worked well. For the sake of consistency, we kept $\mu=0.02$ for \ours as well. We found the rate of client participation to be an important factor for choosing a good value for $\beta$. Therefore, for 1\% client participation experiments we search $\beta$ in $[0.2, 0.4, 0.6]$. For higher rates of client participation, we use the search range of $[0.94, 0.96, 0.98, 1.0]$. For all these cases, $0.96$ and $0.98$ are selected for 10\% and 100\% client participation rates, respectively (both balanced and unbalanced).  
We follow \cite{acar2021federated} for choosing the inference model by averaging client models though the rounds.
Experiments are repeated 5 times, each with a different random seed of data partitioning. The reported mean and standard deviation of the performance are calculated based on the the last round of these 5 instance for each setting.

\subsection{Model Architecture}
We use the same model architectures as \cite{mcmahan2017communication} and \cite{acar2021federated}. For \emnist, the architecture comprises of two fully-connected layers, each with $100$ hidden units. For \cifart{} and \cifarh{}, there are two convolutional layers with $5\times 5$ kernel size and $64$ kernels each, followed by two fully-connected layers with $394$ and $192$ hidden units, respectively. 

\subsection{Baselines}
We compare the performance of \ours against \fedavg \cite{mcmahan2017communication}, \scaffold \cite{karimireddy2020scaffold} and \feddyn \cite{acar2021federated}. These baselines are consistent with the ones that the closest work to us \cite{acar2021federated}, has experimented with\footnote{\feddyn additionally compares with \fedprox \cite{li2020federated}; however, as shown in their benchmarks it performs closer to \fedavg than the other baselines.}. However, we avoided their choice of tuning the number of local epochs since we believe it does not comply with a fair comparison in terms of computation complexity and privacy loss.

\subsection{Evaluation}
\begin{figure}
  \centering
  \def\svgwidth{12.0cm}
  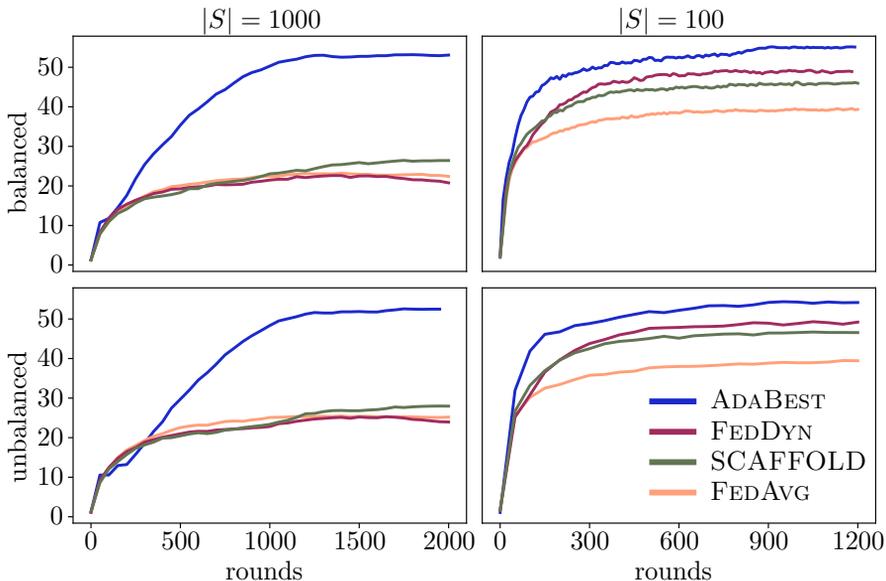
  \caption{Test accuracy on balanced (top) and unbalanced (bottom) settings for training on 1000 (left) and 100 (right) clients. The training dataset is \cifarh{} and $|\gP|$=10}
  \label{fig:scale_robustness}
  
\end{figure}

\tableref{tb:performance-overview} compare the performance of our model to all the baselines in various settings with 100 clients. The results show that our algorithm is effective in all settings. The 1000-device experiments confirm our arguments about the large-scale cross-device setting and practicality of \ours in comparison to the baselines. Our algorithm has notable gain both in the speed of convergence and the generalization performance. This gain is only overtaken for some benchmarks in full client participation settings (CP=100\%) where the best $\beta$ is chosen close to one. According to Remark \ref{re:spectrum}, and the fact that in full participation $\frac{|\gP^t|}{|S^t|}=1$ and $t_i = t'_i + 1$ for all feasible $i$ and $t$, \feddyn and \ours become nearly identical in these settings. In \figref{fig:scale_robustness}, we show the impact of scaling the number of clients in both balanced and imbalanced settings for the same dataset and the same number of clients sampled per round (10 clients). During the hyper-parameter tuning we noticed that the sensitivity of \feddyn and \ours to their $\mu$ is small specially for the cases with larger number of clients.

\begin{table}[tb!]
\begin{center}
    \caption{Mean and standard deviation of test accuracy for various settings. The results are based on 5 random data partitioning seeds. Models are trained for 1k, 1.2k, and 2k rounds, for 1\%, 10\% and 100\% client participation settings, respectively. A \textit{smaller} $\alpha$ indicates \textit{higher} heterogeneity.
    $^*$CP stands for rate of client participation ($\frac{|\gP|}{|S|}$)}
    \label{tb:performance-overview}
        \begin{tabular}{cllcccc}
            \toprule 
            \multicolumn{1}{l}{}  &  & \multicolumn{5}{c}{\textbf{Top-1 Test Accuracy}} \\ \hline $\text{CP}^*$ & {\ \ \ Dataset \ \ \ } & {Setting\ \ } & {\ \ \fedavg \ \ \ } & {\feddyn} & {\scaffold} & {\ \ \ \ours\ \ \ } \\ \hline
            \multirow{9}{*}{1\%}
                & \multirow{3}{*}{\emnist} 
                  & $\alpha$=0.03 & 94.28$\pm$0.07 & 92.42$\pm$0.14 & 93.99$\pm$0.16 & \textbf{94.49$\pm$0.07} \\ 
                & & $\alpha$=0.3  & 94.47$\pm$0.10 & 92.64$\pm$0.31 & 94.34$\pm$0.23 & \textbf{94.72$\pm$0.22} \\  
                & & IID           & 94.04$\pm$1.37 & 92.89$\pm$0.14 & 94.48$\pm$0.11 & \textbf{94.81$\pm$0.08} \\
                \cline{2-7}
                & \multirow{3}{*}{\cifart}          
                  & $\alpha$=0.03 & 78.18$\pm$0.80 & 77.91$\pm$0.79 & 75.83$\pm$2.36 & \textbf{78.44$\pm$1.12} \\  
                & & $\alpha$=0.3  & 82.21$\pm$0.36 & 82.06$\pm$0.17 & 82.96$\pm$0.42 & \textbf{83.09$\pm$0.76} \\  
                & & IID           & 83.84$\pm$0.17 & 83.36$\pm$0.39 & 84.18$\pm$0.26 & \textbf{85.05$\pm$0.31} \\
                \cline{2-7}
                & \multirow{3}{*}{\cifarh}
                  & $\alpha$=0.03 & 47.56$\pm$0.59 & 46.27$\pm$0.65 & 47.29$\pm$0.95 & \textbf{47.91$\pm$0.83} \\
                & & $\alpha$=0.3  & 49.63$\pm$0.47 & 50.53$\pm$0.36 & 52.87$\pm$0.61 & \textbf{53.62$\pm$0.23} \\
                & & IID           & 49.93$\pm$0.36 & 50.85$\pm$0.38 & 53.43$\pm$0.44 & \textbf{55.33$\pm$0.44} \\
            
            \midrule
            \multirow{9}{*}{10\%}
                & \multirow{3}{*}{\emnist}
                  & $\alpha$=0.03 & 93.58$\pm$0.25 & 93.57$\pm$0.20 & 94.29$\pm$0.11 & \textbf{94.62$\pm$0.17} \\ 
                & & $\alpha$=0.3  & 94.04$\pm$0.04 & 93.54$\pm$0.22 & 94.54$\pm$0.11 & \textbf{94.64$\pm$0.11} \\  
                & & IID           & 94.32$\pm$0.10 & 93.60$\pm$0.35 & 94.62$\pm$0.16 & \textbf{94.70$\pm$0.24} \\
                \cline{2-7}
                & \multirow{3}{*}{\cifart}    
                  & $\alpha$=0.03 & 74.04$\pm$0.88 & 76.85$\pm$0.91 & 77.19$\pm$1.10 & \textbf{79.64$\pm$0.58} \\  
                & & $\alpha$=0.3  & 79.74$\pm$0.07 & 81.91$\pm$0.19 & 82.26$\pm$0.38 & \textbf{84.15$\pm$0.36} \\  
                & & IID           & 81.35$\pm$0.23 & 83.56$\pm$0.31 & 83.50$\pm$0.15 & \textbf{85.78$\pm$0.14} \\
                \cline{2-7}
                & \multirow{3}{*}{\cifarh}    
                  & $\alpha$=0.03 & 39.18$\pm$0.56 & 44.24$\pm$0.66 & 45.80$\pm$0.36 & \textbf{48.56$\pm$0.45} \\  
                & & $\alpha$=0.3  & 38.78$\pm$0.35 & 48.92$\pm$0.37 & 46.34$\pm$0.43 & \textbf{54.51$\pm$0.35} \\  
                & & IID           & 37.45$\pm$0.57 & 49.60$\pm$0.24 & 44.30$\pm$0.22 & \textbf{55.58$\pm$0.14} \\
            \midrule
            \multirow{9}{*}{100\%}
                & \multirow{3}{*}{\emnist}
                  & $\alpha$=0.03 & 93.36$\pm$0.15 & 94.18$\pm$0.21 & \textbf{94.38$\pm$0.20} & 94.06$\pm$0.11 \\  
                & & $\alpha$=0.3  & 93.99$\pm$0.19 & 94.23$\pm$0.14 & \textbf{94.53$\pm$0.16} & 94.40$\pm$0.21 \\  
                & & IID           & 94.06$\pm$0.33 & 94.37$\pm$0.15 & 94.63$\pm$0.10 & \textbf{94.69$\pm$0.14} \\
                \cline{2-7}
                & \multirow{3}{*}{\cifart}          
                  & $\alpha$=0.03 & 72.97$\pm$1.09 & \textbf{78.24$\pm$0.77} & 77.64$\pm$0.25 & 78.07$\pm$0.71 \\  
                & & $\alpha$=0.3  & 79.12$\pm$0.15 & 83.19$\pm$0.18 & 82.26$\pm$0.23 & \textbf{83.20$\pm$0.25} \\  
                & & IID           & 80.72$\pm$0.33 & 84.39$\pm$0.20 & 83.55$\pm$0.25 & \textbf{84.75$\pm$0.17} \\
                \cline{2-7}
                & \multirow{3}{*}{\cifarh}
                  & $\alpha$=0.03 & 38.24$\pm$0.63 & 46.00$\pm$0.42 & \textbf{46.51$\pm$0.50} & 46.16$\pm$0.79 \\  
                & & $\alpha$=0.3  & 37.03$\pm$0.35 & 50.42$\pm$0.29 & 45.48$\pm$0.38 & \textbf{50.90$\pm$0.42} \\  
                & & IID           & 35.92$\pm$0.48 & 50.61$\pm$0.25 & 43.73$\pm$0.23 & \textbf{51.33$\pm$0.41} \\
        \end{tabular}
    \end{center}
\end{table}

\label{sec:experiments}

\section{Conclusions}
In this paper, we introduce \ours, an adaptive approach for tackling client drift in Federated Learning. Unlike the existing solutions, our approach is robust to low rates of clients re-sampling, which makes it practical for large-scale cross-device Federated Learning. 
In benchmarks with a large number of clients, our method outperforms the baselines by a wide margin. 
The results show nearly a two-fold improvement in test accuracy over the second-best candidate in some cases.
Our algorithm consumes no more communication bandwidth or storage than the baselines, and it even has a lower compute cost. \ours addresses the instability of norm of gradient estimates used in \feddyn by adapting to the most relevant information about the direction of the client drift. Furthermore, we formulated the general estimate of oracle gradients in a much elegant arithmetic that eliminates the need for the explicit, recursive form used in the previous algorithms.

Future work for this study includes deriving theoretical bounds for our proposed algorithm. Furthermore, in the current paper, the parameter $\beta$ is manually tuned to account for the trade-off between the amount of information retained from the previous estimations of oracle's gradients and the estimation provided by a new round. Developing a method for automatically tuning $\beta$ is an important direction for improving the proposed algorithm.

\smallskip \noindent \textbf{Acknowledgments: } 
The first author wishes to express gratitude for the financial support provided by \emph{MITACS} and \emph{Research Nova Scotia}.
In addition, the fifth author acknowledges Natural Sciences and Engineering research Council of Canada, CHIST-ERA grant CHIST-ERA-19-XAI-0 and the Polish NCN Agency NCN(grant No. 2020/02/Y/ST6/00064). We are grateful to Sai Praneeth Karimireddy, the first author of \cite{karimireddy2020scaffold}, for enlightening us on the proper implementation of \scaffold. William Taylor-Melanson is also acknowledged for reviewing this paper and providing numerous helpful comments.

\clearpage
\bibliographystyle{splncs04}
\bibliography{biblio}

\clearpage
\section*{Appendix}
\setcounter{remark}{0}
\setcounter{theorem}{0}
\setcounter{figure}{3}
\setcounter{table}{2}
\setcounter{algorithm}{1}
\setcounter{section}{0}
\renewcommand{\thesection}{\Alph{section}}
\section{Proofs}

\begin{remark}
Aggregating client models by averaging is equivalent to applying a gradient step of size 1 from the previous round's cloud model using average of client pseudo-gradients or mathematically it is $\bar{\vtheta}^t \leftarrow \frac{1}{|S^t|} \sum_{i\in S^t} \vtheta_i^t = \vtheta^{t-1} - \bar{\vg}^t$.
\end{remark}
\begin{proof}
\begin{equation*}
\begin{split}
    \bar{\vtheta}^t \leftarrow \frac{1}{|S^t|} \sum_{i\in S^t} \vtheta_i^t & = \vtheta^{t-1}- \frac{1}{|S^t|} \sum_{i\in S^t} \vtheta^{t-1} - \vtheta_i^t \\
    & = \vtheta^{t-1}-\frac{1}{|S^t|} \sum_{i\in S^t} \vg_i^t\\ & = \vtheta^{t-1} - \bar{\vg}^t
     \ .
\end{split}
\end{equation*}
\end{proof}


\vspace{0.6cm}
\begin{theorem}
In \feddyn, ${\|\vh^t\|}^2 \leq {\|\vh^{t-1}\|}^2$ requires
\begin{equation*}
\cos( \angle (\vh^{t-1}, \mean{\vg}^{t})) \leq - \frac{|\gP^t|}{2|S^t|} \frac{\|\mean{\vg}^{t}\|}{\|{\vh}^{t-1}\|}
\end{equation*}

\end{theorem}

\begin{proof}
According to \algref{alg:adabest},
\begin{equation*}
 \vh^t \leftarrow \vh^{t-1} + \frac{|\gP^t|}{|S^t|} \mean{\vg}^t. 
\end{equation*}
Applying 2-norm squared on both sides gives
\begin{equation*}
{\|\vh^{t}\|}^2 = {\|\vh^{t-1}\|}^2 + \left(\frac{|\gP^t|}{|S^t|}\right)^{(2)}{\|\mean{\vg}^{t}\|}^2 + 2 \frac{|\gP^t|}{|S^t|} \langle \vh^{t-1}, \mean{\vg}^{t} \rangle.
\end{equation*}
Considering the proposition, we have
\begin{equation*}
    \therefore \ {\|\vh^t\|}^2 \leq {\|\vh^{t-1}\|}^2 \implies \ 2 \langle \vh^{t-1}, \mean{\vg}^{t} \rangle \leq - \frac{|\gP^t|}{|S^t|} {\|\mean{\vg}^{t}\|}^2.
\end{equation*}
with dividing both sides on some positive values, we get
\begin{equation*}
 \frac{\langle \vh^{t-1}, \mean{\vg}^{t} \rangle}{\|\mean{\vg}^{t}\|\|{\vh}^{t-1}\|} \leq - \frac{|\gP^t|}{2|S^t|} \frac{\|\mean{\vg}^{t}\|}{\|{\vh}^{t-1}\|},
\end{equation*}
which is equivalent to the 
\begin{equation*}
\cos( \angle (\vh^{t-1}, \mean{\vg}^{t})) \leq - \frac{|\gP^t|}{2|S^t|} \frac{\|\mean{\vg}^{t}\|}{\|{\vh}^{t-1}\|}.
\end{equation*}

\end{proof}

\vspace{0.6cm}
\begin{remark}
$\mean{\vtheta}^{t-1}-\mean{\vtheta}^{t}$ is equivalent to $\vh^{t-1} + \mean{\vg}^{t}$ in \ours.
\end{remark}
We first add and remove $\vtheta^{t-1}$ from the first side of the equation,
\begin{equation*}
\begin{split}
 \mean{\vtheta}^{t-1}-\mean{\vtheta}^{t} & = \mean{\vtheta}^{t-1}-\mean{\vtheta}^{t} + \vtheta^{t-1} - \vtheta^{t-1}\\ & = (\mean{\vtheta}^{t-1} - \vtheta^{t-1}) + (\vtheta^{t-1} - \mean{\vtheta}^{t} ).
 \end{split}
\end{equation*}
Then, we replace some terms using \eqref{eq:theta} and Remark \ref{re:aggregate} to get
\begin{equation*}
\begin{split}
 \mean{\vtheta}^{t-1}-\mean{\vtheta}^{t} =
 \vh^{t-1} + \mean{\vg}^{t}.
 \end{split}
\end{equation*}

\vspace{0.6cm}
\begin{remark}
Cloud pseudo-gradients of AdaBest form a power series of $\vh^t = \sum_{\tau=0}^{t} \beta^{(t-\tau+1)}\mean{\vg}^{\tau}$, given that superscript in parenthesis means power.
\end{remark}
\begin{proof}
We make the induction hypothesis $\vh^{t-1} = \sum_{\tau=0}^{t-1} \beta^{(t-\tau)}\mean{\vg}^{\tau}$. We need to prove that $\vh^{t} = \sum_{\tau=0}^{t} \beta^{(t-\tau+1)}\mean{\vg}^{\tau}$. 
From \algref{alg:adabest} we have
\begin{equation*}
    \vh^{t} = \beta ( \mean{\vtheta}^{t-1}-\mean{\vtheta}^{t}).
\end{equation*}
Additionally, using Remark \ref{re:gt_rec} it could be rewritten as
\begin{equation*}
    \vh^{t} = \beta ( \vh^{t-1} + \beta \mean{\vg}^{t}).
\end{equation*}
Replacing the induction hypothesis changes it to
\begin{equation*}
\begin{split}
    \vh^{t} &= \beta ( \sum_{\tau=0}^{t-1} \beta^{(t-\tau)}\mean{\vg}^{\tau} + \beta \mean{\vg}^{t}) \\ &= \sum_{\tau=0}^{t-1} \beta^{(t-\tau)}\mean{\vg}^{\tau + 1} + \beta^{(2)} \mean{\vg}^{t} \\ &=\sum_{\tau=0}^{t} \beta^{(t-\tau+1)}\mean{\vg}^{\tau}.
\end{split}
\end{equation*}

\end{proof}

\vspace{0.6cm}

\begin{remark}
\fedavg is a special case of \ours where $\beta=\mu=0$.

\begin{proof}
In \ours, $\mu=0$ makes $\vh_i^t$ zero for all feasible $i$ and $t$. The resulting local update is identical to that of \fedavg. Similarly, $\vh^t$ becomes zero at all rounds if $\beta=0$. So \ours would also have the same server updates as of \fedavg.  
\end{proof}

\vspace{0.6cm}

\end{remark}

\begin{remark}
Server update of \feddyn is a special case of \ours where $\beta=1$ except that \textbf{an extra $\frac{|\gP|}{|S|}$ scalar is applied which also adversely makes \feddyn require prior knowledge about the number of clients}.

\begin{proof}
According to \algref{alg:adabest}, on the server side \ours and \feddyn are different in their update of $\vh^t$. Based on Remark \ref{re:gt_rec}, for $\beta=1$ \ours update is $\vh^t \leftarrow \vh^{t-1} + \mean{\vg}^t$. Comparably the same update in \feddyn is $\vh^t \leftarrow \vh^{t-1} + \frac{|\gP^t|}{|S^t|}\mean{\vg}^t$. Involving $\|S^t\|$ in the update means assuming that prior knowledge on number of total clients is available from the beginning of the training. On the other hand, $\beta=0$ leads to $\vh^t=0$ and consequently the update on the cloud model become $\vtheta^{t} \leftarrow \mean{\vtheta}^{t}$ which is identical to the server update of \fedavg.
\end{proof}

\end{remark}

\vspace{0.6cm}
\begin{theorem}
If $S$ be a fixed set of clients, $\mean{\vtheta}$ does not converge to a stationary point unless $\vh \rightarrow 0$.
\end{theorem}

\begin{proof}
With a minor abuse in our notation for the case of \scaffold/m (the difference only is applying $\vh$ on the clients after $\vtheta$ is sent to them), we can generally state that
\begin{equation*}
    \mean{\vtheta}^{t} \leftarrow \vtheta^{t-1} - \mean{\vg}^{t}= \mean{\vtheta}^{t-1} - (\vh^{t-1}+\mean{\vg}^t).
\end{equation*}
With $S$ being fixed, upon $t \rightarrow \infty$, and convergence of $\mean{\vtheta}$, we expect that $\vg^t \rightarrow \vzero$, so the optimization does not step out of the minima.
In that case. we also expect $\mean{\vtheta}^{t} \approx \mean{\vtheta}^{t-1}$. On the other hand, above formula results in $\mean{\vtheta}^{t} \approx \mean{\vtheta}^{t-1} - \vh^{t-1}$ which holds if $\vh \rightarrow \vzero$.
\end{proof}


\section{Algorithm Notation}

Following \cite{karimireddy2020scaffold} and \cite{acar2021federated}, for the sake of
simplicity and readability, we only presented \algref{alg:adabest} for the balanced
case (in terms of number of data samples per client).
According to \cite{feddynacar}, \scaffold and \feddyn also need the prior
knowledge about the number of clients in order to properly weight their $\vh^t$
accumulation in the case of unbalance data samples.
These weights are used in the form of average samples per client in \cite{feddynacar}.
We eliminate such dependency in our algorithm by progressively calculating this average
throughout the training.
We also confirm that applying the same modification on \scaffold and \feddyn
does not impede their performance nor their stability (experiments on \feddyn and
\scaffold still are done with their original form).
For the experiments, we implemented \scaffold as introduced in the original paper
\cite{karimireddy2020scaffold}; However, for more clarity a modification of it
(\scaffold/m) is contrasted to other methods in \algref{alg:adabest} in which
only the model parameters are sent back to the server (compare it to Algorithm 1 in \cite{karimireddy2020scaffold}).
Note that this difference in presentation is irrelevant to our arguments in this
paper since the more important factor for scalability in the recursion is
$\frac{|S^t|-|\gP^t|}{|S^t|}$.

\section{Algorithmic Costs}
In this section, we compare compute, storage and bandwidth costs of \ours to
that of \fedavg, \scaffold/m and \feddyn.

\subsection{Compute Cost}
Table~\ref{tb:cost_notation} shows the notation we use for formulating the
\textit{costs} of these algorithms.
Algorithm~\ref{alg:adabest_complexity} is an exact repetition of Algorithm
\ref{alg:adabest} except that the compute cost of operations of interest are included
as comments.
These costs are summed in tables Table~\ref{tb:cost_comp_client} and Table
\ref{tb:cost_comp_server} for the client and server sides, respectively.
According to these tables, \ours has lower client-side and server-side compute
costs than \scaffold/m and \feddyn. 

\begin{table}
\caption{Summary of notion used to formulate the algorithm costs}
\label{tb:cost_notation}
\begin{center}
\begin{small}
\begin{tabular}{c@{\hskip 10pt}l}
\toprule
Notation & Meaning \\
\midrule
$n \ $ & Number of parameters of the model $:=|\vtheta|$ \\
$g \ $ & Cost of computing local mini-batch gradients\\
$s \ $ & Cost of summing two floating point numbers\\
$m \ $ & Cost of multiplying two floating point numbers\\
\bottomrule
\end{tabular}
\end{small}
\end{center}
\end{table}

\begin{table}
\caption{Comparing \ours to \fedavg, \scaffold/m, \feddyn in their compute cost of local (client side) operations. See Algorithm~\ref{alg:adabest_complexity} for more detailed comparison}
\label{tb:cost_comp_client}
\begin{center}
\begin{tabular}{l@{\hskip 10pt}l}
\toprule
Algorithm & Client side compute cost \\
\midrule
\fedavg \ \  & $K (g + n s + n m)$ \\
\scaffold/m \ \  & $K (g + n s + n m) + 2 K n s + 2n (s+m) $ \\
\feddyn \ \  & $K (g + n s + n m) + 3 K n s + K n m + n (s+m) $ \\
\ours \ \  & $K (g + n s + n m) + K n s + n (s+m) $ \\
\bottomrule
\end{tabular}
\end{center}
\end{table}

\begin{table}
\caption{Comparing \ours to \fedavg, \scaffold/m, \feddyn in their compute cost of global (server side) operations. See Algorithm~\ref{alg:adabest_complexity} for more detailed comparison}
\label{tb:cost_comp_server}
\begin{center}
\begin{tabular}{l@{\hskip 10pt}l}
\toprule
Algorithm & Server side compute cost \\
\midrule
\fedavg \ \  & $|\gP^t| n s $ \\
\scaffold/m \ \  & $|\gP^t| n s + 2 n s + 2 n m $ \\
\feddyn \ \  & $|\gP^t| n s + 3 n s + n m $ \\
\ours \ \  & $|\gP^t| n s + 2 n s + n m $ \\
\bottomrule
\end{tabular}
\end{center}
\end{table}

\subsection{Storage Cost}
All the three algorithms require the same amount of storage, on the clients and the server. Each client is supposed to store a set of local gradient estimates with size $n$ (see Table~\ref{tb:cost_notation}), noted as $\vh_i^{t}$. Likewise, each algorithm stores the same number of variables on the server so that estimates are forwarded to the next rounds. These variables are introduced with $\vh^t$ in \scaffold/m and \feddyn but with $\mean{\vtheta}^{t-1}$ in \ours. 

\subsection{Communication Cost}
\ours and \feddyn are not different in the way information is communicated between the server and clients; thus, they do not differ in terms of costs of communication bandwidth. That is sending $n$ parameters from server to each selected client at each round and receiving the same amount in the other direction (from each client to the server). The original \scaffold needs doubling the amount of information communicated in each direction ($2 n$). However, \scaffold/m reduces this overhead to 1.5 times (of that of \ours) by avoiding to send the extra variables from clients' to the server (1.5 n).

\begin{algorithm}
   \caption{Compute cost of \colorbox{red!30}{\strut\scaffold/m}, \colorbox{blue!20}{\strut\feddyn}, \colorbox{green!30}{\strut\ours}. Operations that are common among all three algorithms are grayed out. The compute cost of other operations are shown with a comment in front of each line. The variables used to represent the cost of each micro-operation are introduced in \tableref{tb:cost_notation}}
   \label{alg:adabest_complexity}
\begin{spacing}{1.2}
\begin{algorithmic}
   \STATE {\color{gray}{\bfseries Input: $T, \vtheta^0, \mu, \beta$}}
   \FOR{{\color{gray}{$t=1$ {\bfseries to} $T$}}}
   \STATE {\color{gray}~Sample clients $\gP^t \subseteq S^t$.}
   \STATE {\color{gray}~Transmit $\vtheta^{t-1}$ to each client in $\gP^t$}
   \STATE \colorbox{red!30}{\color{gray}Transmit $h^{t-1}$ to each client in $\gP^t$ (\scaffold/m)}
   \FOR{{\color{gray}each client $i \in \gP^t$ {\bfseries in parallel}}}
   \STATE {\color{gray}$\vtheta_i^{t,0} \leftarrow \vtheta^{t-1}$}
   \FOR{{\color{gray}$k=1$ {\bfseries to} $K$}}
   \STATE {\color{gray} Compute mini-batch gradients $L_i(\vtheta_i^{t,k-1})$} \COMMENT{$g$}
   \STATE \colorbox{red!30}{$\vg_i^{t,k-1} \leftarrow \nabla L_i(\vtheta_i^{t,k-1}) - \vh_i^{t'_i} + \vh^t$ (\scaffold/m)} \COMMENT{$2 n s$}
   \STATE \colorbox{blue!20}{$\vg_i^{t,k-1} \leftarrow \nabla L_i(\vtheta_i^{t,k-1}) - \vh_i^{t'_i} - \mu (\vtheta^{t-1} - \vtheta_i^{t, k-1})$ (\feddyn)} \COMMENT{$3 n s + n m$}
   \STATE \colorbox{green!30}{$\vg_i^{t,k-1} \leftarrow \nabla L_i(\vtheta_i^{t,k-1}) - \vh_i^{t'_i}$ (\ours)} \COMMENT{$n s$}
  \STATE {\color{gray}~$\vtheta_i^{t,k} \leftarrow \vtheta_i^{t,k-1} - \eta \vg_i^{t,k-1}$} \COMMENT{$n s + n m$}
   \ENDFOR
   \STATE {\color{gray}~$\vg_i^{t} \leftarrow \vtheta^{t-1} - \vtheta_i^{t,K}$}
    \STATE \colorbox{red!30}{$\vh_i^{t} \leftarrow \vh_i^{t'} - \vh^{t-1} + \frac{1}{K\eta} \vg_i^{t}$
    (\scaffold/m)}\COMMENT{$2 n s + 2 n m$}
   \STATE \colorbox{blue!20}{$\vh_i^{t} \leftarrow \vh_i^{t'_i} + \mu \vg_i^{t}$ (\feddyn)} 
   \colorbox{green!30}{$\vh_i^{t} \leftarrow \frac{1}{t-t'_i}\vh_i^{t'_i} + \mu \vg_i^{t}$ (\ours)} \COMMENT{$n s + n m$}
   \STATE {\color{gray}~$t'_i \leftarrow t$}
   \STATE {\color{gray}~Transmit client model $\vtheta_i^t:=\vtheta_i^{t,K}$}
   \ENDFOR
   \STATE {\color{gray}~$\mean{\vtheta^t} \leftarrow \frac{1}{|\gP^t|}\sum_{i\in \gP^t}\vtheta_i^t$} \COMMENT{$|\gP^t| n s$}
   \STATE \colorbox{red!30}{$\vh^{t} \leftarrow \frac{|S^t|-|\gP^t|}{|S^t|}\vh^{t-1} +\frac{|\gP^t|}{K\eta|S^t|}({\vtheta}^{t-1}-\mean{\vtheta}^t)$ (\scaffold/m)} \COMMENT{$ 2 n s + 2 n m$}
   \STATE \colorbox{blue!20}{$\vh^{t} \leftarrow \vh^{t-1} +\frac{|\gP^t|}{|S^t|}({\vtheta}^{t-1}-\mean{\vtheta}^t)$ (\feddyn)} \COMMENT{$ 2 n s + n m$}
   \STATE \colorbox{green!30}{$\vh^t \leftarrow \beta(\mean{\vtheta}^{t-1}-\mean{\vtheta}^t)$ (\ours)} \COMMENT{$ n s + n m$}
  \STATE \colorbox{red!30} {$\vtheta^{t} \leftarrow \mean{\vtheta}^{t}$ (\scaffold/m)} 
  \STATE \colorbox{blue!20} {$\vtheta^{t} \leftarrow \mean{\vtheta}^{t} - \vh^t$ (\feddyn)} 
  \colorbox{green!30} {$\vtheta^{t} \leftarrow \mean{\vtheta}^{t} - \vh^t$ (\ours)} \COMMENT{$n s$}
   \ENDFOR
\end{algorithmic}
\end{spacing}
\end{algorithm}

More accurate costs requires exact specifications of the system design. For example, the aggregation operation on the server if done in a batch (from a buffer of client delivered parameters) requires more storage but can decrease the compute cost using multi-operand adders on floating-point mantissas such as \emph{Wallace} or \emph{Dadda} tree adders. However, these design choices do not appear to make a difference in the ranking of the costs for algorithms compared in this paper. 

For cost estimates to be more precise, system design specifications must be considered. For instance, using multi-operand adders on floating-point mantissas, such as \emph{Wallace} or \emph{Dadda} tree adders, can reduce the compute cost of the aggregation operation on the server if it is performed in a batch (from a buffer of client-delivered parameters) but requires more storage. However, it does not appear that these design decisions affect the ranking of the costs for the algorithms compared in this paper.

\section{Experiments Details}

\subsection{Evaluation}
There are two major differences between our evaluation and those of prior works. 
\begin{enumerate}
    \item \textbf{We do not consider number of epochs to be a hyper-parameter.}
    Comparing two \fl (FL) algorithms with different number of local epochs,
    is unfair in terms of the amount of local compute costs. 
    Additionally, it makes it difficult to justify the impact of each algorithm on
    preserving the privacy of clients' data.
    This is because the privacy cost is found to be associated with the level of random
    noise in the pseudo-gradients.
    This randomness in turn is impacted by the number of epochs (see page 8 of
    \cite{fowl2021robbing}).
    For an example of comparing algorithms after tuning the number of epochs, refer to
    the Appendix 1 of \cite{acar2021federated} where 20 and 50 local epochs are
    chosen respectively for \fedavg and \feddyn in order to compare their
    performance on \textsc{MNIST} dataset. 
    
    \item \textbf{We consider a hold-out set of clients for hyper-parameter tuning.}
    Although, an {\it on the fly hyper-parameter tuning} is much more appealing in
    FL setting, for the purpose of studying and comparing FL
    algorithms, it is reasonable to consider hyper-parameters are tuned prior to the
    main training phase.
    However, using the performance of the test dataset in order to search for
    hyper-parameters makes the generalization capability of the algorithms that use
    more hyper-parameters questionable.
    Therefore, we set aside a separate set of training clients to tune the
    hyper-parameters for each algorithm individually.
    This may make our reported results on the baselines not exactly matching that of
    their original papers (different size of total training samples).
\end{enumerate}
In addition, we use five distinct random seeds for data partitioning to better justify
our reported performance.
Throughout all the experiments, \sgd with a learning rate of $0.1$ is
used\footnote{We followed \cite{acar2021federated} in choosing this optimization
algorithm and learning rate.} with a round to round decay of $0.998$.
Batch size of $45$ is selected for all datasets and experiments.
Whenever the last batch of each epoch is less than this number, it is capped with
bootstrapping from all local training examples (which can further enhances the privacy
especially for imbalance settings).
We follow \cite{acar2021federated} in data augmentation and transformation.
Local optimization on \cifart{} and \cifarh{} involves random horizontal
flip and cropping on the training samples both with probability of $0.5$.
No data augmentation is applied for experiments on \emnist{}.

\subsection{Implementation}
We used \textsc{PyTorch} to implement our FL simulator.
For data partitioning and implementation of the baseline algorithms, our simulator is
inspired from \cite{feddynacar} which is publicly shared by the authors of
\cite{acar2021federated}.
To further validate the correctness of \scaffold implementation, we consulted the first author of \cite{karimireddy2020scaffold}. Additionally, we cross-checked most of the results our simulator yielded to the ones made by \cite{feddynacar}.

\begin{figure}[tbh!]
  \centering
  \def\svgwidth{12.0cm}
  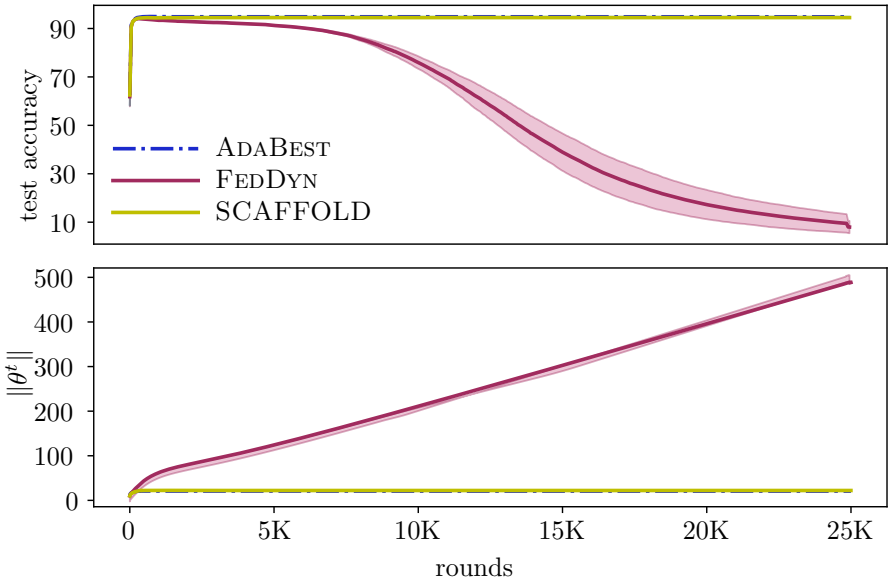
  \caption{Highlighting the instability of \feddyn and its association with the norm of cloud parameters. The training is performed on a comparably easy FL task but for a large number of communication rounds. 
Top and bottom subplots show test accuracy (in percentage) and norm of cloud parameters respectively. The horizontal axis which shows number of communication rounds is shared among subplot}
  \label{fig:stability_emnist}
\end{figure}

\begin{figure}[tbh!]
  \centering
  \def\svgwidth{12.0cm}
  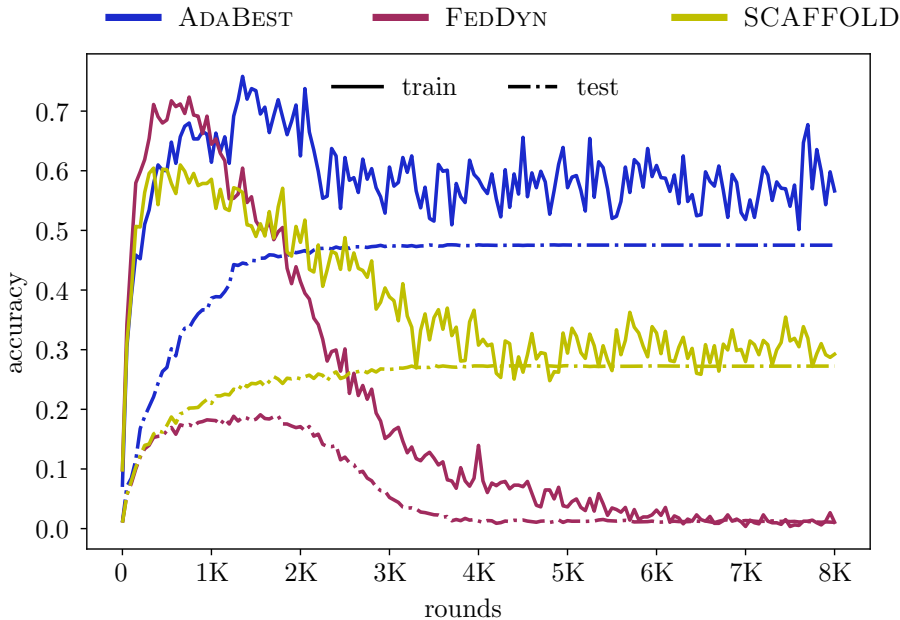
  \caption{Top-1 train and test accuracy score of \ours and the baselines. The solid and dash-dotted lines represent train and test, respectively}
  \label{fig:overfitting}
\end{figure}

\subsection{Stability and $\|\vtheta^t\|$}
\label{sec:stability_sup}
In Figure \ref{fig:stability_c100}, we showed an experimental case of low client participation to demonstrate how $\|\vtheta^t\|$ is associated with instability of \feddyn. In this experiment, the training split of \cifarh{} is partitioned over $1100$ clients from which $1000$ is used for training. The partitioning is balanced in terms of number of examples per client and the labels are skewed according to our $\alpha=0.3$ heterogeneity setup (see Section \ref{sec:experiments} for detailed explanation). In each round, $5$ clients are drawn uniformly at random. This low client participation rate is more likely to occur in a large-scale  (in terms of number of clients) cross-device FL compared to the setting used for reporting the performances in Table~\ref{tb:performance-overview} and most of those of our prior works. In Figure \ref{fig:stability_emnist} we repeat the same experiment; except that a much simpler FL task is defined. In this experiment the training split of \emnist{} dataset is partitioned into 110 clients, 100 of which are used for training. Partitions are IID (labels are not skewed). Even though this task is much simpler than the previous one, still \feddyn fails to converge when the training continues for a large number of rounds.

\subsection{Overfitting Analysis}
It is important not to confuse \feddyn's source of instability with overfitting. To confirm this, we can compare the average train and test accuracy of the same rounds while the model is being trained. Figure \ref{fig:overfitting} compares \scaffold, \feddyn, and \ours in this manner. The configuration used in this experiment is identical to the configuration of the experiment in Figure \ref{fig:stability_c100}, with the exception that it corresponds to a single data partitioning random seed for clarity. The train accuracy is calculated by averaging the train accuracy of participating clients. The results suggest that train accuracy of the baselines is severely declined as the training continues while \ours is much more stable in this regard.

\subsection{$\mu$-sensitivity}

\begin{figure}[tbh!]
  \centering
  \def\svgwidth{12.0cm}
  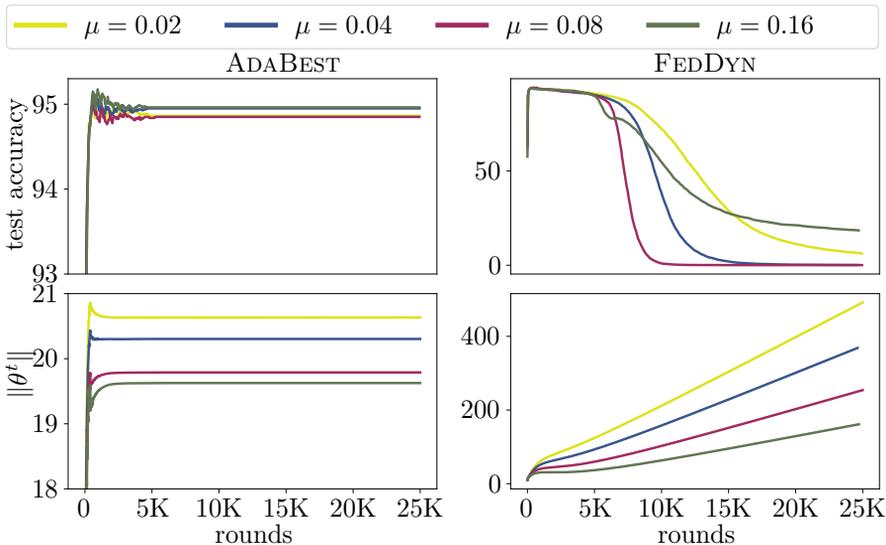
  \caption{$\mu$ sensitivity for \ours (on the left) and \feddyn (on the right). The horizontal axis which shows the number of communication rounds is shared for each column of subplots. Top and bottom row show the test accuracy (in percentage) and norm of cloud parameters respectively. Note that the vertical axis is not shared as for clarity. This means that the scale of difference between converging point of $\|\vtheta^t\|$ in \ours is largely different from the divergence scale of the same quantity for \feddyn}
  \label{fig:mu_sensitivity}
\end{figure}

During the hyper-parameter tuning phase, we only tune $\beta$ of \ours and set its $\mu$ constantly to $0.02$. This is done throughout all the experiments except the experiment presented in this section which especially investigates the sensitivity of \ours to varying $\mu$. Therefore, in practice we have not treated $\mu$ as a hyper-parameter but rather chose the value that works best for \feddyn. For this experiment, we use the same setup as the one presented in Section \ref{sec:stability_sup} on \emnist{} dataset. The only difference is that we vary $\mu$ in the range of $\{0.02\times 2^{(k)}\}_{k=1}^{k=3}$. Figure \ref{fig:mu_sensitivity} depicts the outcome of this experiment along with the result of the same analysis on \feddyn so it would be easy to compare the sensitivity of each of these algorithms to their local factor $\mu$. As suggested by the top left subplot in this Figure, \ours can even achieve higher numbers in terms of the test accuracy than what is reported in \tableref{tb:performance-overview}. The scales on the vertical axis of the subplots related to \ours (on the left column) are zoomed-in to better show the stability of our algorithm throughout the training. On the other hand, the performance and stability of \feddyn shows to be heavily relied on the choice of $\mu$ when the training continues for a large number of rounds.

\subsection{$\beta$-sensitivity}

\begin{figure}[tb!]
  \centering
  \def\svgwidth{12.0cm}
  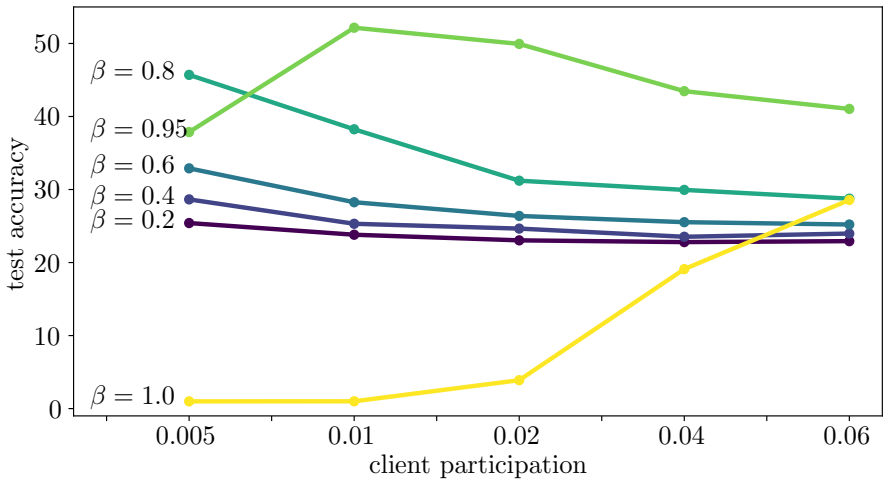
  \caption{The sensitivity of the test accuracy (in percentage) for different rates of client participation and values of $\beta$. The training is done on a partitioning of \cifarh{} with $1000$ training clients. The numbers on the horizontal axis show the fractions of the total clients sampled at each round}
  \label{fig:beta_sensitivity}
\end{figure}

To investigate how the choice of $\beta$ impacts the generalization performance, we conduct an experiment with varying $\beta$ and the rate of client participation. The relation comes from the fact that in lower rates of client participation, the variance of the pseudo-gradients is higher and so a lower $\beta$ is required both in order to avoid explosion of $\|\vtheta^t\|$ and also to propagate estimation error from the previous rounds as explained in \ref{sec:adaptability}. For this experiment we use the same setup as the one used in Figure \ref{fig:stability_c100} (see Section \ref{sec:stability_sup} for details). Figure \ref{fig:beta_sensitivity} implies that when the rate of client participation is $0.005$ ($5$ clients participating in each round out of $1000$ training clients) the optimal $\beta$ is between $0.8$ and $0.95$. With a larger rate of client participation, the optimal value moves away from $0.8$ towards between $0.95$ and $1.0$;

This observation is aligned with our hypothesis on the impact of variance of pseudo-gradients on the estimation error and norm of the cloud parameters. Another interesting point is that for a wide range of $\beta$ values, the performance remains almost stable regardless of the rate of client participation. Additionally, $\beta$ values closer to 1 seem to be suitable for higher rates of client participation, which is suggested by the curve corresponding to $\beta=1.0$ rising as the rate of client participation increases. This case ($\beta=1.0$) is the most similar to formulation of our baselines where the estimations of oracle gradients are not scaled properly from one round to another.

\end{document}